\documentclass[10pt,a4paper,twoside]{article}

\usepackage[utf8]{inputenc}
\usepackage[T1]{fontenc}
\usepackage{graphicx}
\usepackage{lipsum}
\usepackage{hyperref}
\usepackage{amsmath}
\usepackage{talnrecital2013}
\usepackage[frenchb]{babel}

\DeclareMathOperator*{\argmax}{arg\,max}

\title{Systèmes du LIA à DEFT'13}

\author{Xavier Bost\up{1}\quad Ilaria Brunetti\up{1,6}  \quad Luis Adri\'an Cabrera-Diego\up{1} \\ \quad Jean-Valère Cossu\up{1} \quad Andréa Linhares\up{1,5} \quad Mohamed Morchid\up{1}\\  \quad  Juan-Manuel Torres-Moreno\up{1,2,3,4} \quad Marc El-Bèze\up{1,2,4} \quad Richard Dufour\up{1,4} \\
{\small  (1) LIA, 339, chemin des Meinajariès  84912 Avignon Cedex 09\\ 
  (2) SFR Agorantic Université d'Avignon et des Pays de Vaucluse, 84000 Avignon Cedex\\
  (3) \'Ecole Polytechnique de Montréal, 2900 Bd Edouard-Montpetit  Montréal, QC H3T1J4 \\ 
  (4) Brain \& Language Research Institute, 5 avenue Pasteur, 13604 Aix-en-Provence Cedex 1\\
  (5) Universidade Federal do Ceara Rua Estanislau Frota, S/N, CEP 62.010-560 – Sobral/ CE Brésil\\
  (6) Equipe Maestro INRIA, 2004, Route des Lucioles, 06902 Sophia Antipolis\\
  \texttt{prenom.nom@univ-avignon.fr} \\ 
}}

\begin{document}

\maketitle

\resume{
La campagne Défi de Fouille de Textes (DEFT) en 2013 s'est intéressée à deux types de fonctions d'analyse du langage, la classification de documents et l'extraction d'information dans le domaine de spécialité des recettes de cuisine. 
Nous présentons les systèmes du LIA appliqués à DEFT 2013.
Malgré la difficulté des tâches proposées, des résultats intéressants ont été obtenus par nos systèmes.\\
}

\abstract{The Systems of LIA at DEFT'13}{
The 2013 {\sl Défi de Fouille de Textes} (DEFT) campaign is interested in two types of language analysis tasks, the document classification and the information extraction in the specialized domain of cuisine recipes.
We present the systems that the LIA has used in DEFT 2013.
Our systems show interesting results, even though the complexity of the proposed tasks.\\
}

\motsClefs{Classification de textes; Extraction d'information; Méthodes statistiques; Domaine de spécialité}
{Text Classification; Information Extraction; Statistical Methods; Speciality domain}


\section{Description des tâches et des données}

Le Défi Fouille de Textes (DEFT) s'intéresse en 2013 à deux types de fonction d'analyse du langage, la classification de documents (tâches 1 à 3) et l'extraction d'information (tâche 4), ceci dans un domaine de spécialité :
les recettes de cuisine. 
Pour cette nouvelle édition du défi\footnote{\url{http://deft.limsi.fr/2013/}}, 
l'équipe du Laboratoire Informatique d'Avignon (LIA) a décidé de participer aux tâches suivantes :

\begin{itemize}
{\sl
    \item T1. L'identification à partir du titre et du texte de la recette de son niveau de difficulté sur une échelle à 4 niveaux : très facile, facile, moyennement difficile, difficile ;
    \item T2. L'identification à partir du titre et du texte de la recette du type de plat préparé : entrée, plat principal, dessert ;
    \item T4. L'extraction à partir du titre et du texte d'une recette de la liste de ses ingrédients.
}
\end{itemize}

	Le corpus, fourni par DEFT au format {\it XML}, est composé de 13 684 recettes.
	Nous avons choisi d'augmenter notre base de données en extrayant des recettes de cuisine à partir de 4 sites spécialisés\,: Cuisine  AZ (\url{http://www.cuisineaz.fr}), Madame Le Figaro (\url{http://madame.lefigaro.fr/recettes}), 750 grammes (\url{http://www.750g.com/}) et Ptit Chef (\url{http://www.ptitchef.com/}). 
Environ 50k recettes supplémentaires ont pu être obtenues à partir de ces sites.
Notons cependant que les recettes téléchargées ne comportaient pas toujours les mêmes champs d'information que ceux contenus dans les recettes fournies par les organisateurs de la campagne.

\section{Méthodes utilisées}

Pour répondre à ce défi, nous avons employé plusieurs méthodes de classification (modèles probabilistes discriminants, {\sl boosting} de caractéristiques textuelles) et d'extraction d'information, ainsi que leurs combinaisons.

\subsection{Modèles discriminants}

Le système DISCOMP\_LIA a été appliqué aux tâches 1 et 2. Son fonctionnement repose essentiellement sur les modèles discriminants décrits dans \cite{Torres:11}.
Ces modèles servent aussi bien à agglutiner des mots afin de produire des termes composites ayant un fort pouvoir discriminant (sans toutefois passer en dessous d'un seuil préfixé de couverture), qu'à pondérer de façon plus appropriée ces termes dans le calcul de l'indice de similarité (ici Cosine revisitée) entre chaque recette et le sac de mots de chacune des classes (3 ou 4 selon les tâches).

Trois nouveautés ont été introduites pour prendre en compte les particularités du défi :
\begin{itemize}
	\item nous avons opté pour une approche hiérarchique propre à chacune des 2 tâches :
\begin{itemize}
\item pour la tâche T1 : identification du niveau {\it facile} ou {\it difficile} d'une recette, puis dans chacun des 2 groupes, différenciation en 2 sous-groupes ({\it moyennement} ou {\it très difficile}) ;
\item pour la tâche T2 : identification du type {\it dessert} ou {\it autre} ; dans le cas d'une identification en type {\it autre}, différenciation entre {\it plat principal} et {\it entrée} ; 
\end{itemize}
	\item nous avons appliqué le même système avec un paramétrage {\sl ad hoc} à 2 jeux de données : le titre de la recette seul pour le premier et pour le second l'ensemble formé par le contenu et le titre de la recette. Une combinaison linéaire entre les scores des 2 systèmes a permis d'obtenir une nouvelle distribution de probabilités sur les hypothèses de classes ;
	\item la formation par agglutination des termes et ce, surtout dans le titre, a fait apparaître des sous-types de recettes (quiches, soupes, gâteaux) qui ont permis fournir aux modèles de meilleurs points d'appui pour identifier la classe visée. Dans les cas où l'indice de pureté de {\it Gini} valait 1, et où la couverture était élevée, nous avons enrichi le sac de mots avec des composants factices de façon à compenser le bruit qui pouvait être engendré par le reste de la recette. Ceci a été fait de façon semi-automatique, mais devrait pouvoir être complètement automatisé.
\end{itemize}

Notons par ailleurs que nous nous sommes servis du critère de pureté de {\it Gini} pour sélectionner quelques mots erronés ce qui, après correction,
a permis d'augmenter leur couverture sans altérer leur pouvoir discriminant.

\subsection{Algorithmes de {\sl boosting}}
\label{sec:boosting}

Pour les tâches de classification (1 et 2), nous proposons une manière de combiner différentes caractéristiques textuelles et numériques au moyen d'un algorithme d'apprentissage automatique : le {\it Boosting}~\cite{Schapire03}. Son objectif principal est d'améliorer (de \og booster  \fg) la précision de n'importe quel algorithme d'apprentissage permettant d'associer, à une série d'exemples, leur classe correspondante. Le principe général du \textit{Boosting} est assez simple : la combinaison pondérée d'un ensemble de classifieurs binaires (appelés classifieurs \textit{faibles}), chacun associé à une règle différente de classification très simple et peu efficace, permettant au final d'obtenir une classification robuste et très précise (classifieur \textit{fort}). La seule contrainte de ces classifieurs \textit{faibles} est d'obtenir des performances meilleures que le hasard. Dans le cas des classifieurs binaires, l'objectif est de classifier plus de 50~\% des données correctement.

Nous proposons d'utiliser l'algorithme {\it AdaBoost} car il présente de nombreux avantages dans le cadre des tâches de classification proposées. En effet, l'algorithme est très rapide et simple à programmer, applicable à de nombreux domaines, adaptable aux problèmes multi-classes. Il fonctionne sur le principe du \textit{Boosting}, où un classifieur \textit{faible} est obtenu à chaque itération de l'algorithme \textit{AdaBoost}. Chaque exemple d'apprentissage possède un poids, chaque tour de classification permettant de les re-pondérer selon le classifieur \textit{faible} utilisé. Le poids d'un exemple bien catégorisé est diminué, au contraire d'un exemple mal catégorisé, 
pour lequel on augmente son poids. 
L'itération suivante se focalisera ainsi sur les exemples les plus \og difficiles  \fg~ (poids les plus élevés). L'algorithme a été implémenté dans l'outil de classification à large-marge \textit{BoosTexter}~\cite{Schapire00}, permettant de fournir comme caractéristiques au classifieur des données numériques mais également des données textuelles. 
Les classifieurs \textit{faibles} sur les données textuelles peuvent prendre en compte des $n$-grammes de mots. 
Nous avons utilisé l'outil {\it IcsiBoost}~\cite{Favre07}, qui est l'implémentation open-source de cet outil. {\it IcsiBoost} a notamment l'avantage de pouvoir fournir, pour chaque exemple à classifier, un score de confiance pour chacune des classes entre 0 (peu confiant) et 1 (très confiant).

\subsubsection{Niveau de difficulté}
\label{sec:difficulte}

Pour l'approche utilisant l'algorithme de classification {\it AdaBoost}, nous avons retenu différentes caractéristiques qui ont été fournies en entrée pour l'apprentissage du classifieur :

\begin{itemize}
\item données textuelles
\begin{enumerate}
\item $n$-grammes de mots du titre de la recette avec taille maximum de 3 mots ;
\item $n$-grammes de mots du texte de la recette avec taille maximum de 4 mots ;
\item uni-gramme sur la liste des ingrédients obtenue automatiquement (voir section \ref{sec:ingredients}).
\end{enumerate}
\item données numériques continues
\begin{enumerate}
\item nombre de mots du titre de la recette ;
\item nombre de mots du texte de la recette ;
\item nombre de phrases du texte de recette ;
\item nombre de séparateurs du texte de la recette (point, virgule, deux-points, etc.) ;
\item taille de la liste des ingrédients obtenue automatiquement.
\end{enumerate}
\end{itemize}

\`A la fin de ce processus d'apprentissage, la liste des classifieurs sélectionnés est obtenue tout comme le poids de chacun d'eux, afin d'utiliser les exemples les plus discriminants pour chaque classe (chaque niveau de difficulté est considéré dans cette tâche comme une classe). Nous avons composé un sous-corpus à partir du corpus d'apprentissage fourni par les organisateurs de la tâche afin d'optimiser le nombre de classifieurs {\it faibles} nécessaires. 
Ce sous-corpus est composé de 3~863 recettes respectant la distribution des classes, le reste étant utilisé pour l'apprentissage (environ 72\% des données d'apprentissage). Notons que tout le corpus d'apprentissage sera néanmois utilisé pour entraîner les classifieurs pendant la phase de test de la campagne. Enfin, une attention particulière a été portée sur la normalisation des données textuelles du titre et de la description des recettes. En effet, les données textuelles \og brutes  \fg~   fournies par les organisateurs possédaient un vocabulaire très bruité  principalement à cause des nombreuses abbréviations (par exemple \og th  \fg~   pour \og thermostat  \fg) et fautes d'othographe (par exemple \og échalote  \fg~  et \og échalotte  \fg). Quatre traitements particuliers ont été appliqués sur le texte :

\begin{enumerate}
\item suppression de la ponctuation et isolation de chaque mot (exemple : \og et couper l'oignon.  \fg~   devient \og et$\sqcup$couper$\sqcup$l'$\sqcup$oignon  \fg) ;
\item utilisation d'une liste manuelle de normalisation de mots (exemple : \og kg  \fg~   devient \og kilogramme  \fg) ;
\item conversion automatique des chiffres en lettres ;
\item regroupement des $n$-grammes de mots les plus fréquents au sein d'une même entité au moyen de l'outil {\it lia\_tagg}\footnote{http://lia.univ-avignon.fr/fileadmin/documents/Users/Intranet/chercheurs/bechet/download\_fred.html} (exemple : \og il y a  \fg~   est considéré comme un mot \og il\_y\_a  \fg).
\end{enumerate}

Ce texte normalisé est utilisé par l'algorithme {\it AdaBoost} pour la recherche des classifieurs {\it faibles} sur les données textuelles lors de la phase d'apprentissage, mais également pour la recherche du niveau de difficulté associé à une recette lors de la phase de test. Au final, nous obtenons pour chaque recette à classifier, un score de confiance pour chaque niveau de difficulté, permettant de choisir le niveau de difficulté de la recette (score le plus élevé).

\subsubsection{Type de plat}

Dans ce problème de classification au moyen de l'approche par {\it Boosting}, nous avons utilisé les mêmes caractéristiques et normalisations que celles décrites dans la tâche de recherche du niveau de difficulté (voir section~\ref{sec:difficulte}). Pour chaque recette, nous obtenons donc un score de confiance pour chaque type de plat qui pourra être utilisé pour choisir la classe à associer à une recette.

\subsection{SVMs}

Le corpus des recettes $\mathbb{X}$ étant donné, il s'agit d'élaborer
à partir d'un sous-ensemble $\mathbb{D} \subset \mathbb{X}$ de
recettes annotées par classe (niveau de difficulté ou type de plat
selon la tâche), une méthode de classification $\gamma$ qui à toute
recette associe une classe. En notant $\mathbb{C}$ l'ensemble des
classes, nous avons donc~:
\begin{equation}
\gamma : \mathbb{X} \longrightarrow \mathbb{C}
\end{equation}

La troisième des méthodes appliquées a consisté, pour chaque couple de
classes $(c, c') \in \mathbb{C} \times \mathbb{C}$ (avec $c \not =
c'$), à apprendre et à appliquer un classifieur binaire
$\gamma_{(c,c')} : \mathbb{X} \rightarrow \{c, c'\}$ à base de SVMs.

En notant $s_{(c,c')}(r)$ le score obtenu selon le classifieur
$\gamma_{(c,c')}$ par la recette $r \in \mathbb{X}$, nous avons alors, avec
$c' \in \mathbb{C}$~:
\begin{equation}
\gamma(r) = \argmax_{c \in \mathbb{C}} \left
(s_{(c, c')}(r) \right )
\end{equation}

Lors des phases d'apprentissage et d'application des classifieurs
binaires, les recettes ont été représentées vectoriellement dans
l'espace du lexique de référence. Le poids $w(t)$ du terme $t$ d'une recette
$r$ dans le vecteur associé est donné par~:
\begin{equation}
w(t) = tf(t).idf(t) = tf(t).\log \left ( \frac{|\mathbb{X}|}{df(t)} \right )
\end{equation}
où $tf(t)$ désigne le nombre d'occurrences du terme $t$ dans la
recette et $df(t)$ le nombre de recettes du corpus $\mathbb{X}$ où le
terme $t$ apparaît.
L'approche décrite a été appliquée pour les deux tâches de classification.

Dans le cas de la prédiction du type de plat, le lexique a cependant
été filtré avant vectorisation des recettes sur le critère de
l'information mutuelle entre un terme $t$ et une classe $c \in
\mathbb{C}$ \cite{Manning:2008:IIR:1394399}.
Après sélection des termes, seuls les 10 000 termes les plus porteurs
d'information sur les classes ont alors été retenus.

\subsection{Similarité cosinus}

La quatrième approche, seulement utilisée pour la prédiction du type
de plat, repose elle aussi sur une représentation vectorielle des
documents~: à chaque recette $r$, nous faisons correspondre un vecteur
$v_r$ dont les composantes $w_r(t)$ dans l'espace du lexique sont
données, pour chaque terme $t$ présent dans le corps de la recette,
par~:

\begin{equation}
\begin{aligned}
w_r(t) & = tf(t).idf(t).G(t) \\
& = tf(t).idf(t).\sum_{c \in \mathbb{C}} \mathbb{P}^2(c|t) \\
& = tf(t).idf(t).\sum_{c \in \mathbb{C}} \left ( \frac{df_c(t)}{df_{\mathbb{T}}(t)} \right ) ^2
\end{aligned}
\end{equation}
où $G(t)$ désigne l'indice de {\it Gini}, $df_{\mathbb{T}}(t)$ est le
nombre de recettes du corpus d'apprentissage $\mathbb{T} \subset
\mathbb{X}$ contenant le terme $t$ et $df_c(t)$ correspond au nombre
de recettes du corpus d'apprentissage annotées selon le type $c \in
\mathbb{C}$.

A chaque classe $c \in \mathbb{C}$, nous faisons par ailleurs correspondre
un vecteur dont les composantes $w_c(t)$ sont données dans l'espace du
lexique par~:
\begin{equation}
w_c(t) = df_c(t).idf(t).G(t)
\end{equation}
où $df_c(t)$ désigne le nombre de recettes annotées selon le type de
plat $c$ où le terme $t$ apparaît.

Une mesure de similarité $s(r, c)$ entre une recette $r \in
\mathbb{X}$ et un type de plat $c \in \mathbb{C}$ est alors donnée par
le cosinus de l'angle formé par les vecteurs $v_r$ et $v_c$~:
\begin{equation}
s(r, c) = \cos(\widehat{v_r,v_c})
= \frac{\sum_{t \in r \cap c}w_r(t).w_c(t)}{\sqrt{\sum_t w_r(t)^2.w_c(t)^2}}
\end{equation}

Dans ces conditions la classe $\gamma(c)$ attribuée à une recette $r$
est celle qui maximise la mesure de similarité cosine~:
\begin{equation}
\gamma(r) = \argmax_{c \in \mathbb{C}} \left ( s(r, c) \right )
\end{equation}

Le vocabulaire utilisé lors de la phase de vectorisation était
constitué des seuls termes $t$ dont l'indice de Gini $G(t)$ était au
moins égal à 0.45.

Ce seuil a été fixé empiriquement par maximisation du macro F-score
sur un corpus de développement constitué de 3~864 des 13~864 recettes du
corpus d'apprentissage.

\subsection{Fusion des systèmes}

\label{sec:fusion}

Pour chacune des tâches de classification, plusieurs méthodes ont donc
été appliquées~: trois pour la tâche~1, quatre pour la tâche~2.

Pour chaque couple $(r,c) \in \mathbb{X} \times \mathbb{C}$ associant
une recette du corpus à une classe, nous disposions donc d'un score
$s_i(r,c)$ pour chacune des méthodes de classification (avec $i = 1,
..., 3$ ou $i = 1, ..., 4$ selon la tâche).

\subsubsection{Normalisation des résultats par méthode}

Les scores produits en sortie des différents systèmes de
classification ont d'abord été normalisés de manière à ce que la somme
des prédictions sur l'ensemble des classes pour une méthode donnée
soit égale à 1. En notant $n_i(r,c)$ le score normalisé obtenu selon
la $i-$ème méthode pour le couple $(r,c) \in \mathbb{X} \times
\mathbb{C}$, nous avons donc~:
\begin{equation}
n_i(r,c) = \frac{s_i(r,c)}{\sum_{j=1}^{|\mathbb{C}|} s_i(r,c_j)}
\end{equation}
où $c_j$ désigne la $j-$ème classe ($j = 1, ..., 4$ ou $j = 1, ...3$
selon la tâche de classification).

Deux méthodes de fusion ont été appliquées après normalisation des
scores $s_i(r,c)$.

\subsubsection{Combinaison linéaire des scores}

La première méthode de fusion a consisté, pour une recette donnée $r$,
à départager les classes candidates par combinaison linéaire des
scores obtenus selon les différentes méthodes de classification. En
notant $\gamma(r)$ la classe conjecturée, on a donc~:
\begin{equation}
\gamma(r) = \argmax_{c \in \mathbb{C}} \left ( \sum_{i=1}^{m} n_i(r,c) \right )
\end{equation}
où $m$ correspond au nombre de méthodes appliquées pour la tâche considérée.

\subsubsection{Méthode \textsc{Electre}}

Issue du domaine de l'optimisation multi-objectif discrète, la méthode
\textsc{Electre}~\cite{BR91} consiste à définir sur l'ensemble $\mathbb{C}$ des
classes candidates une relation $\mathcal{S} \subset \mathbb{C} \times
\mathbb{C}$ dite de surclassement. De manière informelle, nous pouvons dire qu'une
classe $c$ surclasse une classe $c'$ si elle la domine sur un nombre
\og{}important \fg~ de méthodes et si, sur les éventuelles méthodes
restantes où elle est dominée par $c'$, elle ne l'est pas
au-delà d'un certain seuil fixé pour chaque méthode. Plus formellement~:
\begin{equation}
c \mathcal{S} c' \Longleftrightarrow
\begin{cases}
& conc(c,c') \geqslant sc \\
& v(c,c') = 0
\end{cases}
\end{equation}
où $conc(c,c') \in [0,1]$ désigne l'indice de concordance entre les
classes candidates $c$ et $c'$, $sc \in [0,1]$ un seuil dit de
concordance fixé empiriquement et $v(c,c') \in \{0,1\}$ un indice
binaire dit {\it de veto}.

L'indice de concordance $conc(c,c')$ évalue le taux de méthodes
(éventuellement pondérées) selon lesquelles $c$ domine $c'$~:
\begin{equation}
conc(c,c') = \frac{\sum_{i \in M(c,c')} p_i}{\sum_{i \in M} p_i}
\end{equation}
où $M$ désigne l'ensemble des méthodes, $M(c,c')$ l'ensemble des
méthodes sur lesquelles $c$ domine (non strictement) $c'$ et $p_i$ le
poids accordé à chaque méthode $i \in M$.

L'indice binaire $v(c,c')$ (à valeurs dans l'ensemble $\{0,1\}$) fixe
la valeur du veto de $c'$ vers $c$ selon les modalités suivantes (en
notant $n_i(r,c)$ le score obtenu selon la $i$-ème méthode par le
couple $(r,c)$ associant une recette à une classe)~:
\begin{equation}
v(c,c') =
\begin{cases}
& 1 \Longleftrightarrow \exists i \in M : n_i(r,c') > n_i(r,c) \text{ et }
n_i(r,c') - n_i(r,c) \geqslant v_i \\
& 0 \text{ sinon}
\end{cases}
\end{equation}
où $v(i) \in [0,1]$ est la valeur de veto associée à la $i-$ème
méthode.

Les valeurs des différents paramètres ont été fixées empiriquement à
partir du corpus de développement par maximisation des métriques
appliquées pour évaluer la classification.

Les poids $p_i$ associés aux différentes méthodes ont tous été fixés à
$p_i = 1$. Le seuil de concordance a été fixé à $sc = 0,7$ pour la
tâche~1 et $sc = 0,6$ pour la tâche~2~; enfin, la valeur de veto $v_i$ a été
fixée à $v_i = 0,5$ pour chacune des méthodes et quelle que soit la
tâche considérée.

Le noyau de la relation $\mathcal{S} \subset \mathbb{C} \times
\mathbb{C}$ est alors constitué des éventuelles classes candidates non
surclassées.

Dans le cas d'un noyau {\it singleton}, la classe conjecturée pour la recette est l'unique classe élément du noyau.

Dans le cas de noyau vide ou de noyau formé de plusieurs classes concurrentes, c'est la meilleure classe candidate selon la méthode de
combinaison linéaire qui a été retenue.

\subsection{Extraction d'ingrédients}
\label{sec:ingredients}

La tâche 4 est une tâche pilote d'extraction d'information.
La difficulté est multiple, car les auteurs des recettes n'utilisent pas tous les ingrédients, ou ils ont introduit librement
des ingrédients non listés. Par exemple, il n'est pas rare d'avoir des passages comme \og mélanger énergiquement tous les ingrédients \fg~
(recette 27174), qui ne permettent pas d'extraire directement les ingrédients.

Nous avons attaqué ce problème en utilisant deux approches : l'une à base de règles et l'autre probabiliste.
L'idée de base pour les règles a été la suivante:
soit $A$ l'ensemble de mots de la recette $i$; soit $B$ l'ensemble de mots de la liste DEFT (références d'apprentissage ou \textsl{gold-standard} d'évaluation).
Éliminer tous les mots de $A$ qui ne sont pas dans $B$. 
Cela produit une liste d'ingrédients candidats $L_c$.

Cette liste peut contenir des termes génériques comme \og VIANDE \fg, \og FROMAGE \fg~ou \og POISSON \fg, présents dans le texte de la recette.
Afin de mieux identifier ce terme, nous avons employé une approche probabiliste naïve.
Nous avons calculé la probabilité d'avoir un certain type de viande $x$, étant donné une liste d'ingrédients $L_c=(L_1,L_2,...,L_n)$:
\begin{equation}
   p(x|L_c) = P(x \cap L_c)/p(L_c) ; \quad x = \{\textrm{VIANDE, FROMAGE, POISSON}\}
\end{equation}
Les termes ainsi identifiés, sont alors injectés dans la liste extraite $L_c$, ce qui produit la liste deifnitive $L$.

La liste d'ingrédients $L$ ainsi obtenue a été utilisée dans les systèmes de {\sl boosting} (c.f. section\ref{sec:boosting}) dans les tâches T1 et T2.

\section{Résultats et discussion}

Les systèmes ont été évalués en utilisant les mesures proposées par DEFT :
le F-score pour les tâches 1 et 2, et la mesure {\sl Mean Average Precision} (MAP) de TREC pour la tâche d'extraction.

\subsection{Tâches de classification}

Pour la tâche de classification en niveau de difficulté (T1), nos
trois runs étaient donnés par~:

\begin{itemize}
  \item Run~1~: SVMs seuls ;
  \item Run~2~: fusion des trois méthodes par \textsc{Electre} ;
  \item Run~3~: combinaison linéaire des trois méthodes.

\end{itemize}

Pour la tâche de classification en type de plat (T2)~:

\begin{itemize}
  \item Run~1~: Modèle discriminant seul ;
  \item Run~2~: fusion des quatre méthodes par \textsc{Electre} ;
  \item Run~3~: combinaison linéaire des quatre méthodes.
\end{itemize}

Les résultats obtenus sont récapitulés dans les
tableaux~\ref{table_res_T1} (tâche 1) et~\ref{table_res_T2} (tâche
2).

Pour la tâche~1 et pour chacun des trois runs, nous fournissons, outre les
F-scores, la distance moyenne (micro-écart) de l'hypothèse à la
référence sur l'échelle des quatre niveaux de difficulté, deux niveaux
contigüs étant distants de 1.

\begin{table} [h]
\begin{center}
  \begin{tabular}{|c|c|c|c|}
    \hline
    \textsc{T1}& \textbf{run~1} & \textbf{run~2} & \textbf{run~3} \\
    \hline
    \textbf{Micro F-score} & \textbf{0.5916} & 0.5812 & 0.5877 \\
    \hline
    \textbf{Macro F-score} & 0.4531 & 0.4531 & \textbf{0.4569} \\
    \hline
    \textbf{Distance moyenne} & \textbf{0.4409} & 0.4586 & 0.4465 \\
    \hline
  \end{tabular}
\end{center}
\caption{\label{table_res_T1} {\it Résultats obtenus sur le corpus de
    test~-~Tâche~1}}
\end{table}

\begin{table} [h]
\begin{center}
\begin{tabular}{|c|c|c|c|}
    \hline
    \textsc{T2} & \textbf{run~1} & \textbf{run~2} & \textbf{run~3} \\
    \hline
    \textbf{Micro F-score} & 0.8760 & 0.8860 & \textbf{0.8886} \\
    \hline
    \textbf{Macro F-score} & 0.8706 & 0.8795 & \textbf{0.8824} \\
    \hline
\end{tabular}
\end{center}
\caption{\label{table_res_T2} {\it Résultats obtenus sur le corpus de
    test~-~Tâche~2}}
\end{table}

Pour la tâche~2, la combinaison des méthodes permet d'obtenir des
scores systématiquement supérieurs à ceux que nous obtenons en appliquant
une méthode isolée.

Pour la tâche d'évaluation du niveau de difficulté, les bénéfices
retirés de l'utilisation d'une méthode de fusion sont moins nets et
apparaissent dépendants de la métrique appliquée lors de l'évaluation.

Ceci a permis au LIA de se positionner dans la tâche 1 : 3ème/6
et dans la tâche 2 : 1er/5.

En ce qui concerne la tâche T2, nous croyons que l'évaluation aurait
dû prendre en compte la dimension muti-labels de plusieurs recettes.
Par exemple la recette 18~052 (présente dans le test) se termine par
 \og servir tiède ( aussi bon chaud que froid ou réchauffé), en
entrée ou en plat \fg~est annotée uniquement comme une entrée.  Si un
système produit l'étiquette {\sl plat principal}, faut-il pour autant
compter cela comme une erreur ?

Relevons aussi l’aspect subjectif de la tâche T1. 
En effet, 
dire qu’une recette est facile ou difficile sans tenir compte qui en est l’auteur c’est faire fi de son niveau d’expertise ! ! !

Nous émettons également l'idée que les macro-mesures devraient être privilégiées aux micro-mesures car ces dernières
pousseraient à mettre en \oe uvre des stratégies négligeant les
classes minoritaires. Nous pensons donc que cette façon d'évaluer serait plus adaptée à une utilisation orientée application.

\subsection{Tâche d'extraction}

L'évaluation de cette tâche est assez subjective, malgré son caractère d'extraction d'information.
Par exemple,  dans l'ensemble d'apprentissage, la recette 10~514 dit : \og 40 cl d'Apremont ou autre blanc de Savoie sec (facultatif, mais donne plus de goût)  \fg. Les références DEFT indiquent {\bf mais} comme ingrédient de la tartiflette.
Evidemment nous n'avons pas incorporé l'information des ingrédients d'apprentissage, à la différence de la méthode DEFT ayant généré le {\sl gold-standard}.

Nous avons obtenu un score de MAP=0.6364 en mesure {\it qrel} dans le corpus d'apprentissage, et dans le classement officiel un score 
de MAP=0.6287 dans le corpus de test. Cela a positionné l'équipe du LIA au rang 3ème/5. 
Bien que standard, la mesure MAP pourrait avoir intégré les accents dans l'évaluation des ingrédients extraits :
cela aurait permis de lever des ambiguïtés qui ont été inutilement introduites par la désaccentuation.

Nous pensons que les références fournies par DEFT dans l'ensemble d'apprentissage, ont contribué à rendre la tâche floue.
En effet, cela n'a pas de sens d'extraire {\sl papier sulfurisé} ou {\sl couteau} (extraite à partir de {\sl pointe de couteau}) comme ingrédients.
De même, l'ingrédient {\it maïs}, incorrectement extrait des recettes d'apprentissage comportant l'expression {\sl mais...}, 
n'est pas apparu dans les recettes de test comportant ladite expression.

\subsection{Comparaison versus les humains}

Nous avons mis en place une petite expérience impliquant des êtres humains.
Pour la tâche T4, nous avons demandé à 7 annotateurs (experts ou non en matière culinaire), d'effectuer la tâche DEFT.
Ceci nous a permis de positionner nos systèmes par rapport aux performances des personnes. 
Puisque les personnes ont des connaissances extra-linguistiques, les juges humains ont eu comme seule consigne
celle de ne pas consulter des ressources externes (par exemple des ressources électroniques ou des livres de cuisine) pour classer les recettes ou pour extraire les ingrédients.
Le tableau \ref{tab:humains} montre la moyenne MAP pour chaque annotateur $A_i$ sur 50 recettes de la tâche T4, choisies au hasard.
La moyenne générale pour les personnes est de MAP=0.5433 et pour le système $S$ de MAP={\bf 0.6570}. 
Ceci montre que dans ce sous-ensemble, nos systèmes sont au-dessus des performances atteintes par les humains.

\begin{table}[h!]
\centering
\begin{tabular}{|c|c|c|c|c|c|c||c|}
  \hline
  A1 & A2 & A3 & A4 & A5 & A6 & A7 & $S$\\
  \hline
  0.5333 & 0.6029 & 0.5271 & 0.5880 & 0.5313 & 0.4679 & 0.5529 & \bf 0.6570\\
  \hline
\end{tabular}
\caption{Performance des personnes sur la tâche T4.}
\label{tab:humains}
\end{table}

Il est clair que, même les humains ayant une expertise culinaire, ont obtenu de piètres notes dans
cette tâche.
 Cela s'explique par la mauvaise qualité du {\sl gold standard} fourni par DEFT. Par exemple, la crème ou le café (cuillère à café, recette 90~806) semblent subir une extraction de nature aléatoire... 
D'ailleurs, nous avons souvent été surpris par
les ingrédients de référence de DEFT utilisés dans telle ou telle recette, 
qui ne peuvent en aucun cas avoir été employés de façon conjointe. 
Par exemple, dans la recette 64~761 (dessert), 
la référence DEFT liste parmi les ingrédients  \og moules \fg~ et \og caramel\fg~! Évidemment il s'agit de l'ustensile {\sl moule à charlotte}.
Malheureusement il s’agit de problèmes récurrents. Sans parler de l’eau ou de la pâte (déclinée
comme pâté, pâtes fraîches, brisée, sablée, etc.) qui semblent être détectées aléatoirement. Pour répondre au
défi, nous avons développé plusieurs systèmes d’extraction d’ingrédients qui ont dû être modifiés (souvent à l'encontre de ce qui nous semblait logique) afin de rendre
une sortie proche de celle des références. 
Ceci revient en fin de compte à modéliser la machine d’extraction de DEFT. 
Les références étant de qualité discutable, est-il intéressant de chercher à reproduire la sortie d'une telle machine ? 
\section{Conclusions}

Malgré la quantité d'erreurs présentes dans le corpus d'apprentissage, nos algorithmes ont conduit à des résultats intéressants.
Nous voulons ajouter quelques commentaires par rapport à ce défi.
Nous avons observé que l'auto-évaluation du niveau de difficulté des recettes par ceux qui les publient est
un exercice très subjectif, assez souvent sujet à caution. En ce qui concerne le type de plat, l'évolution du
mode de vie rend assez floue la distinction entre plat principal et hors d'\oe uvre.
Les tartes salées, les quiches ou les tartines sont de plus en plus considérées comme des plats \og de résistance  \fg.
Ceci conduit souvent les internautes à exprimer de façon explicite leur indécision quant à la catégorie à retenir.
Ces observations n'ont pas suffi à atténuer notre perplexité lorsque nous avons constaté que certaines méthodes
à base de \og cuisine  \fg~   algorithmique réussissaient à faire mieux que des approches sophistiquées.
Pour autant, nous n'avons pas cédé à la tentation de les retenir pour en faire une soumission !

\section*{Remerciements} 

Nous remercions les annotateurs qui ont bien voulu nous aider dans cette tâche ! 
Patricia Vel\'azquez, Sulan Wong,  Mariana Tello, Alejandro Molina et Grégoire Moreau.
Nous tenons aussi à remercier les organisateurs de la campagne d'évaluation, sans qui nous n'aurions pu participer à ce défi. 


\bibliographystyle{apalike}
\bibliography{biblio}

\begin{thebibliography}{}

\bibitem[Favre et~al., 2007]{Favre07}
Favre, B., Hakkani-T{\"u}r, D., and Cuendet, S. (2007).
\newblock {Icsiboost}.
\newblock \url{http://code.google.com/p/icsiboost}.

\bibitem[Manning et~al., 2008]{Manning:2008:IIR:1394399}
Manning, C.~D., Raghavan, P., and Sch\"{u}tze, H. (2008).
\newblock {\em Introduction to Information Retrieval}.
\newblock Cambridge University Press, New York, NY, USA.

\bibitem[Roy, 1991]{BR91}
Roy, B. (1991).
\newblock The outranking approach and the foundations of {ELECTRE} methods.
\newblock {\em Theory and Decision}, 31:49--73.

\bibitem[Schapire, 2003]{Schapire03}
Schapire, R.~E. (2003).
\newblock {The Boosting Approach to Machine Learning: An Overview}.
\newblock In {\em Nonlinear Estimation and Classification}.

\bibitem[Schapire and Singer, 2000]{Schapire00}
Schapire, R.~E. and Singer, Y. (2000).
\newblock Boos{T}exter: A boosting-based system for text categorization.
\newblock In {\em Machine Learning}, volume~39, pages 135--168.

\bibitem[Torres-Moreno et~al., 2011]{Torres:11}
Torres-Moreno, J., El-Bèze, M., Bellot, P., and F., B. (2011).
\newblock {\em {Peut-on voir la détection d’opinions comme un problème de
  classification thématique ?}}, chapter~9.
\newblock Hermes Lavoisier.

\end{thebibliography}

\end{document}